\documentclass[10pt,twocolumn,letterpaper]{article}

\usepackage{iccv2023AuthorKit/iccv}
\usepackage{times}
\usepackage{epsfig}
\usepackage{graphicx}
\usepackage{amsmath}
\usepackage{amssymb}

\usepackage{multirow} 
\usepackage{pifont} 
\usepackage{subcaption} 
\usepackage{xcolor} 

\usepackage[pagebackref=true,breaklinks=true,letterpaper=true,colorlinks,bookmarks=false]{hyperref}

\iccvfinalcopy 

\def\iccvPaperID{10565} 

\begin{document}

\title{MapFormer: Boosting Change Detection by Using Pre-change Information}

\author{Maximilian Bernhard, Niklas Strauß, Matthias Schubert\\
LMU Munich, MCML\\
{\tt\small \{bernhard,strauss,schubert\}@dbs.ifi.lmu.de}
}

\maketitle

\begin{abstract}
Change detection in remote sensing imagery is essential for a variety of applications such as urban planning, disaster management, and climate research. However, existing methods for identifying semantically changed areas overlook the availability of semantic information in the form of existing maps describing features of the earth's surface.
In this paper, we leverage this information for change detection in bi-temporal images. We show that the simple integration of the additional information via concatenation of latent representations suffices to significantly outperform state-of-the-art change detection methods.
Motivated by this observation, we propose the new task of \emph{Conditional Change Detection}, where pre-change semantic information is used as input next to bi-temporal images. 
To fully exploit the extra information, we propose \emph{MapFormer}, a novel architecture based on a multi-modal feature fusion module that allows for feature processing conditioned on the available semantic information. We further employ a supervised, cross-modal contrastive loss to guide the learning of visual representations. Our approach outperforms existing change detection methods by an absolute 11.7\% and 18.4\% in terms of binary change IoU on DynamicEarthNet and HRSCD, respectively. Furthermore, we demonstrate the robustness of our approach to the quality of the pre-change semantic information and the absence pre-change imagery.
The code is available at \url{https://github.com/mxbh/mapformer}.
\end{abstract}

\begin{figure}
    \centering
    \includegraphics[width=\columnwidth]{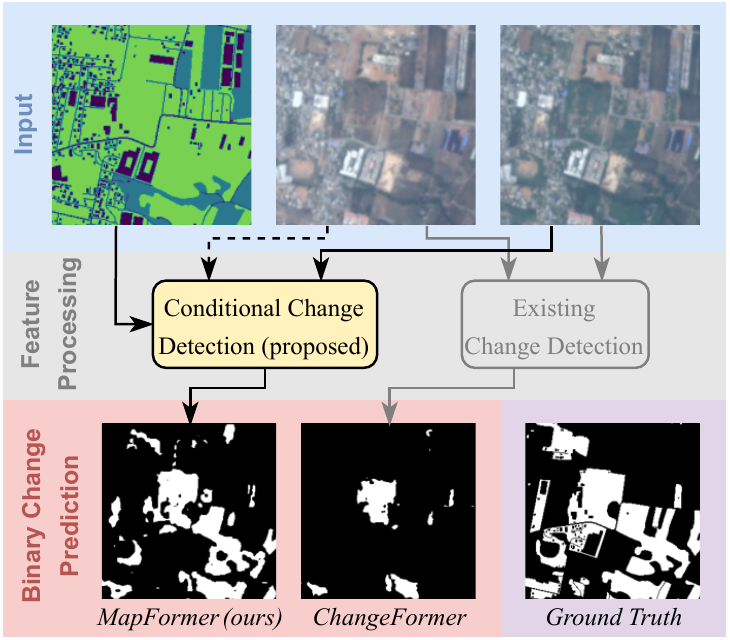}
    \caption{Advantage of our approach over existing change detection methods (represented by ChangeFormer~\cite{changeformer} here) on a sample of DynamicEarthNet~\cite{dynearthnet}. Employing semantic pre-change information allows to extract better features and improves change detection quality.}
    \label{fig:teaser}
\end{figure}
\section{Introduction}

Earth observation data has become increasingly available in recent years, providing valuable insights into various fields such as climate research, disaster management, and urban planning. In the course of this, enormous efforts have been made towards creating rich semantic maps of the earth. 
For instance, OpenStreetMap provides vast amounts of data containing manual annotations around the globe. In addition, deep learning techniques are nowadays used to produce semantic map data at a large scale, e.g.,~\cite{africa-buildings} generated building footprints for the entire African continent.
As a result, we live in a world that has been extensively mapped. 
However, we live in a constantly changing world where the earth's surface is subject to various influences. These influences can be natural, e.g., earthquakes, extreme weather, floods, wildfires, and changes in vegetation, but also directly caused by humans, e.g., building construction, deforestation, and agriculture.
In fact, the impact of civilization on our planet has never been higher than in recent years, with drastic consequences for ecosystems and wildlife~\cite{human-footprint,earth-transformed}.
Therefore, monitoring changes in the earth's surface with up-to-date earth observation data is becoming increasingly important.

In technical terms, this challenge is addressed by the task of change detection, where bi-temporal images are compared in order to segment pixels that have semantically changed. Today, deep-learning-based methods for change detection achieve state-of-the-art performance~\cite{cd-survey-mdpi,cd-survey-ieee}. However, the research community so far mostly ignored the fact that existing semantic information like maps may be employed for change detection. 
To the best of our knowledge, semantic map information has been considered an additional input next to bi-temporal images for detecting change only for the specific task of updating road networks~\cite{beyond-road-extraction}. 
Although this approach is only possible if semantic information for the area of interest is available, we argue that maps are available for most parts of the world today. Alternatively, pre-trained neural networks can be used to generate a map containing the features of interest. Furthermore, when updating a map, a pre-change version of the map has to be available in the first place.

In this paper, we tap into this direction in a general way and establish the novel task of \emph{Conditional Change Detection} (see Figure~\ref{fig:teaser}). 
We demonstrate that the usage of semantic information in form of a pixelwise map has a strong impact on change detection by showing that a simple baseline (concatenating bi-temporal image features and map features) is already able to outperform state-of-the-art change detection methods, which do not use semantic pre-change information. 
Furthermore, we propose \emph{MapFormer}, a novel architecture to fully exploit the additional information. At its core, we integrate a newly designed multi-modal feature fusion module based on the idea that the pre-change semantics should dynamically influence the processing of hidden features. Additionally, we apply a supervised contrastive loss on the learned image features, guiding the image encoder during training and further improving the performance. 
We also show that our approach can be applied without bi-temporal imagery, detecting change by only processing a pre-change map and current imagery. We call this task \emph{Cross-modal Change Detection}.
We conduct experiments on the publicly available and challenging datasets DynamicEarthNet~\cite{dynearthnet} and HRSCD~\cite{hrscd}. 

To sum up, our main contributions are as follows:
\begin{itemize}
    \setlength\itemsep{-0.5em}
    \item We investigate the novel tasks of \emph{Conditional Change Detection} and \emph{Cross-modal Change Detection} and introduce simple baselines that outperform state-of-the-art change detection methods.
    \item We propose \emph{MapFormer}, a novel architecture consisting of a multi-modal feature fusion module and a supervised contrastive loss, allowing us to effectively merge image features and semantic information.
    \item We conduct detailed experiments and ablations demonstrating our approach's consistent performance gains, robustness, and practicability.
\end{itemize}

\section{Related Work}
Change detection in remote sensing images has attracted much attention over the last years, resulting in numerous publications in this field~\cite{cd-survey-mdpi,cd-survey-ieee,fhd,changeformer,changer,snunet,change-is-everywhere,unsupervised-cd,selfpair}.
Most state-of-the-art methods employ deep learning and introduce various sophisticated ways of merging bi-temporal image features.
For instance, \cite{fc-siam} proposes a series of fully convolutional architectures for change detection that rely on different fusion mechanisms for the bi-temporal image features. Furthermore, \cite{snunet} uses NestedUNet~\cite{nestedunet} in combination with channel attention.
With the introduction of transformer architectures into computer vision ~\cite{vit,detr,segformer}, recent architectures for change detection also utilize attention blocks. 
BiT~\cite{bit} was the first of these methods, combining a CNN backbone with cross-attention layers in the decoder head. Subsequently, works like CDViT~\cite{cdvit}, ChangeFormer~\cite{changeformer}, FHD~\cite{fhd}, and Changer~\cite{changer} use the attention mechanism in various forms, pushing the state-of-the-art in bi-temporal change detection even further. Our general architecture is inspired by ChangeFormer and FHD, which adopt the MiT backbone and the MLP head from SegFormer~\cite{segformer} and add specialized mechanisms to fuse the bi-temporal image features.

LEVIR-CD~\cite{levir}, DSIFN-CD~\cite{dsifn}, WHU~\cite{whu}, and S2Looking~\cite{s2looking} are common benchmark datasets for bi-temporal change detection, where networks are provided with co-registered, bi-temporal images as input and have to predict a pixelwise, binary change mask. Recently, a new dataset for change detection has been published with DynamicEarthNet~\cite{dynearthnet}, which surpasses previous datasets in terms of volume and diversity. In addition, models have to predict semantic segmentation masks containing the land cover classes next to the binary change masks for this dataset. 
However, none of these methods or datasets consider inputs other than bi- or multi-temporal images. Thus, previous approaches ignore the fact that pre-change semantic information for the area of interest is often available.
For semantic segmentation on the other hand, it has been shown that semantic maps extracted from geographic information systems (GIS) can be effectively used to improve segmentation quality~\cite{segmentation-osm,building-footprint-gis}.
In a change detection setting, only for the specific task of road network extraction \cite{beyond-road-extraction} challenged the community to come up with solutions for updating road networks when pre-change road networks are given. 
Consequently, their approach considers graph data as input and output.
In contrast, we consider the pixelwise change detection task when pre-change semantic information in the form of a pixelwise map is available and show that this paradigm opens up exciting new possibilities. 


\section{Conditional Change Detection}
\begin{table}
\centering
\begin{tabular}{@{\hspace{0.1em}}l@{\hspace{0.1em}}|@{\hspace{0.3em}}c@{\hspace{0.1em}}c@{\hspace{0.1em}}c@{\hspace{0.1em}}|c@{\hspace{0.1em}}c@{\hspace{0em}}}
\hline 
\multirow{2}{*}{\hspace{2cm}\textbf{Task}} & \multicolumn{3}{c|}{\textbf{Input}} & \multicolumn{2}{c@{\hspace{0.1em}}}{\textbf{Output}}  \\
 & $I^{(1)}$ & $I^{(2)}$ & $m^{(1)}$ & $\hat{b}$ & $\hat{m}^{(2)}$ \\ \hline
Bi-temp.~Bin.~Change~Det.~(BCD)              & \ding{51} & \ding{51} &           & \ding{51} &           \\
Bi-temp.~Sem.~Change~Det.~(SCD)              & \ding{51} & \ding{51} &           & \ding{51} & \ding{51} \\ 
\hspace{1em}\text{[new]\hspace{1.12em} Conditional BCD} & \ding{51} & \ding{51} & \ding{51} & \ding{51} &  \\
\hspace{1em}\text{[new]\hspace{1.2em} Conditional SCD} & \ding{51} & \ding{51} & \ding{51} & \ding{51} & \ding{51} \\
\hspace{1em}\text{[new]\hspace{1em} Cross-modal BCD}                         &           & \ding{51} & \ding{51} & \ding{51} &  \\
\hspace{1em}\text{[new]\hspace{1.1em} Cross-modal SCD}                         &           & \ding{51} & \ding{51} & \ding{51} & \ding{51} \\ \hline
\end{tabular}
\caption{Inputs and outputs for different change detection tasks.}
\label{tab:tasks}
\end{table}

We consider two points in time $t_1$ (pre-change) and $t_2$ (post-change). Correspondingly, we have two image versions $I^{(1)}$ and $I^{(2)}$ for the same geographic location at $t_1$ and $t_2$, respectively. The ground-truth binary mask $b$ indicates which pixels in the images have changed semantically between $t_1$ and $t_2$. Furthermore, we consider semantic maps $m^{(1)}$ and $m^{(2)}$ containing the semantic class $c \in \mathcal{C}$ for each pixel in $I^{(1)}$ and $I^{(2)}$, respectively. Thus, $b$ can be obtained from $m^{(1)}$ and $m^{(2)}$ via a pixelwise inequality operation.
%
As we propose new tasks in this paper (\emph{Conditional Change Detection} and \emph{Cross-modal Change Detection}), we provide a differentiated description of the existing and novel tasks here. An overview can be seen in Table~\ref{tab:tasks}. 

\textbf{Existing Tasks}
In general, change detection tasks differ in two aspects: input and output. We can distinguish w.r.t.~the output between binary and semantic change detection (BCD and SCD) for every combination of inputs. In BCD, the goal is to predict the binary change mask $\hat{b}$, indicating which pixels have changed semantically. SCD was recently proposed in~\cite{dynearthnet} and constitutes an extension of BCD. Here, not only the binary change mask $\hat{b}$ is predicted, but also the semantic segmentation $\hat{m}^{(2)}$ of the post-change image $I^{(2)}$. This allows us to tell which regions have changed and how they have changed. 
The corresponding evaluation metric $SCS$ proposed in~\cite{dynearthnet} combines the binary change IoU $BC$ and the semantic change metric $SC$ as their arithmetic mean. The semantic change metric $SC$ is the standard mIoU from semantic segmentation, except that it is computed only on the pixels that have changed, i.e. 
\begin{align}
    &SC = \frac{1}{|\mathcal{C}|} \sum_{c \in \mathcal{C}} 
    \frac{|\{b = 1\} \cap \{m^{(2)}=c)\} \cap \{\hat{m}^{(2)} = c \}|}
         {|\{b = 1\} \cap (\{m^{(2)}=c)\} \cup \{\hat{m}^{(2)} = c \})|}, \nonumber \\
         &BC = IoU(b, \hat{b}) \quad \text{and} \quad SCS = \frac{BC + SC}{2}
\end{align}

Regarding the input, the existing settings consider bi-temporal images $I^{(1)}$ and $I^{(2)}$. Thus, we will refer to the existing tasks as bi-temporal BCD and bi-temporal SCD.

\textbf{Conditional Change Detection}
While the aforementioned change detection tasks assume bi-temporal images as input, semantic information in the form of a segmentation map $m^{(1)}$ at $t_1$ may be employed additionally. 
That is, given pre- and post-change images together with pre-change semantic information $m^{(1)}$, the goal is to predict the binary change mask $\hat{b}$ for Conditional BCD (as well as the semantic segmentation map $\hat{m}^{(2)}$ for Conditional SCD). 

\textbf{Cross-modal Change Detection}
When pre-change semantic maps are available, change detection can also be conducted without bi-temporal, but solely uni-temporal imagery. In this case, only comparing the semantic information $m^{(1)}$ and the current imagery $I^{(2)}$ can suffice to detect semantic changes. In other words, this setting corresponds to Conditional Change Detection with uni-temporal images. We call this task \textit{Cross-modal Change Detection} as the change has to be detected across different modalities (semantic pre-change map and post-change image).

\section{MapFormer}
\subsection{Overall Architecture}
\begin{figure*}
    \centering
    \includegraphics[width=0.9\textwidth]{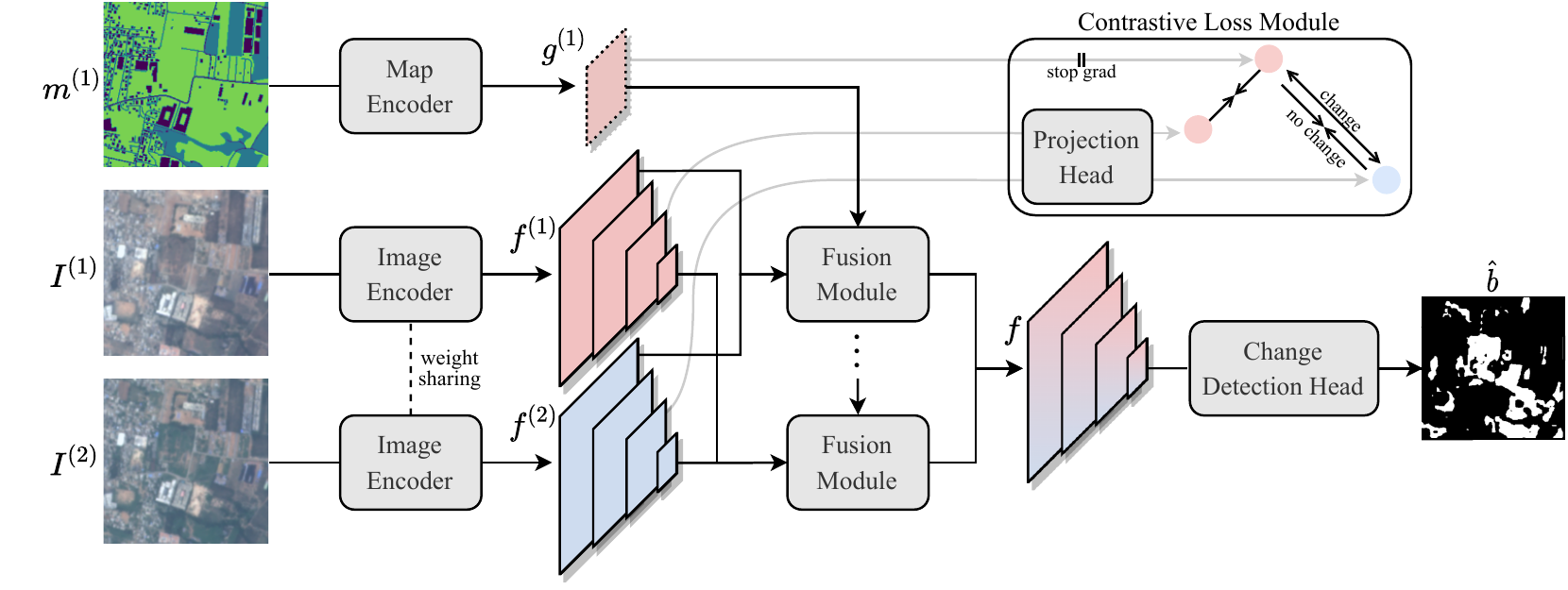}
    \caption{\textbf{The proposed MapFormer framework} consists of two main components: a novel fusion module, which combines inputs from different points in time and modalities, and a contrastive loss module, which is only used during training.}
    \label{fig:arch}
\end{figure*}
To handle these new tasks, we propose a framework, which is inspired by~\cite{fhd,changeformer} (see Figure~\ref{fig:arch}). We feed the images $I^{(1)}$ and $I^{(2)}$ through the same backbone to obtain multi-scale feature maps $f^{(1)}$ and $f^{(2)}$. For each scale, we have a separate fusion module (see Section~\ref{sec:fusion}) that handles the features from $f^{(1)}$ and $f^{(2)}$ at the corresponding scale. In addition to the image features, in our setting, we also feed semantic information in the form of an encoded map $g^{(1)}$ into the fusion modules. These map features are obtained from $m^{(1)}$ via a shallow encoder (see Section~\ref{sec:arch-details}). The outputs of the fusion modules are then multi-scale features $f$, containing information of both points in time $t_1$ and $t_2$. Thus, we can use them in a prediction head to output the binary change map $\hat{b}$. In terms of architecture, this prediction head is a standard semantic segmentation head~\cite{segformer}. Furthermore, we apply a contrastive loss on the image features $f^{(1)}$ and $f^{(2)}$, which helps the model to learn semantically more meaningful representations (see Section~\ref{sec:contrastive-loss}).

For many applications, it may be interesting to determine into which semantic class a pixel has changed. In this case, we also need to derive a semantic segmentation map $\hat{m}^{(2)}$ for $I^{(2)}$. We can accomplish this either by feeding the uni-temporal features $f^{(2)}$ into a separate head for semantic segmentation or by predicting both $\hat{b}$ and $\hat{m}^{(2)}$ based on the merged features $f$. 
We analyze both variants in Section~\ref{sec:scd}. Training a semantic segmentation head on uni-temporal image features also allows applying the trained model without pre-change information as we can use the predicted semantic segmentation $\hat{m}^{(1)}$ instead of $m^{(1)}$. Results for this setting are provided in Section~\ref{sec:bcd}.
Furthermore, for Cross-modal Change Detection, the design of our fusion module (described next) enables us to omit the image features $f^{(1)}$ from $I^{(1)}$. We evaluate this setting in Section~\ref{sec:cross-modal-bcd}.

\subsection{Multi-modal Feature Fusion}
\label{sec:fusion}
\begin{figure}
    \centering
    \includegraphics[width=\columnwidth]{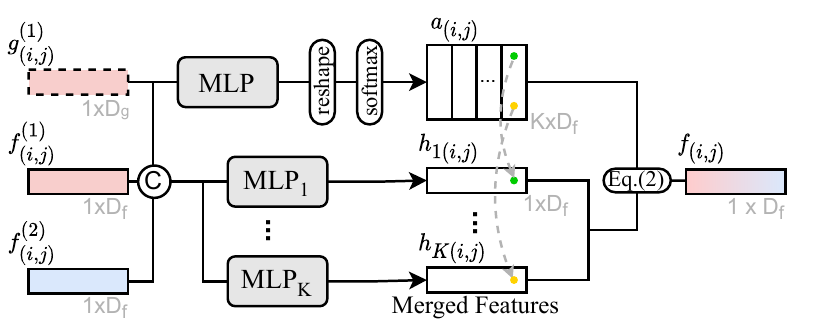}
    \caption{\textbf{Our multi-modal feature fusion module} for fusing bi-temporal image features and semantic input features.}
    \label{fig:fusion}
\end{figure}

The main challenge in bi-temporal change detection is to design an effective method for fusing the bi-temporal image features. In the case of Conditional (and Cross-modal) Change Detection, we additionally have to bring together different modalities, i.e., the image features and the available semantic information. To this end, we propose a novel multi-modal feature fusion module.
Subsequently, we follow~\cite{fhd,segformer} and assume that the spatial context is encoded within the backbone, so we only consider pointwise operations, treating each spatial location separately.

A standard way of merging multi-modal features is a simple concatenation along the channel dimension. However, we argue that the semantic information in our setting can be used more effectively if we use it to modulate the features dynamically. Thus, the given information should influence the processing of image features and thereby improve perception.
Such data-dependent feature processing can be achieved by dynamically predicting the weights of one or more layers from the features themselves~\cite{dynamicmlp,condinst}. While such techniques are powerful, they come at a significant computational cost. In fact, applying a method like DynamicMLP~\cite{dynamicmlp} to a segmentation task like ours by predicting dynamic weights for each pixel separately is intractable without drastically reducing the number of channels (or the spatial resolution).
Therefore, we resort to a restricted attention-style approach. More precisely, to keep the computational cost tractable, we restrict the attention to a set of $K$ values for each output dimension, thereby greatly reducing the complexity compared to unrestricted attention.

A visualization of our proposed multi-modal feature fusion module can be seen in Figure~\ref{fig:fusion}. First, we concatenate the bi-temporal features $f^{(1)}_{(i,j)}$ and $f^{(2)}_{(i,j)}$ and the map feature $g^{(1)}_{(i,j)}$ for each spatial location $(i,j)$. This concatenated feature is fed through $K$ separate MLPs (implemented with grouped, pointwise convolutions) to obtain $K$ joint representations $(h_{k(i,j)})_{k=1...K}$ with $D_f$ channels. Here, $D_f$ refers to the dimensionality of $f^{(t)}_{(i,j)}$, $t=1,2$, and $K$ is a hyperparameter for tuning the semantic variety. 
These $K$ joint representations $(h_{k(i,j)})_{k=1...K}$ can be interpreted as views, encoding different aspects of the same spatial location.
To associate the map information $g^{(1)}_{(i,j)}$ with these views, we employ a channelwise attention mechanism.
In particular, we compute attention weights by feeding the map features $g^{(1)}_{(i,j)}$ into an additional MLP. The result of shape $K\times D_f$ is softmaxed over the $K$-dimension providing a matrix of attention scores with entries $a_{(i,j,k,d)}$ for location $(i,j)$. Thus, each channel of the module output $f_{(i,j)}$ separately attends to the $K$ joint representations $(h_{k(i,j)})_{k=1...K}$. In mathematical terms, we compute
\begin{equation}
f_{(i,j)} = \left(\sum_{k=1...K} a_{(i,j,k,d)} h_{k(i,j,d)} \right)_{d=1...D_f}
\end{equation}

The merged features $f_{(i,j)}$ are then collected for the different scales of the backbone feature maps and passed to the prediction head.

\subsection{Contrastive Loss}
\label{sec:contrastive-loss}
For unchanged areas, we want $f^{(1)}$, $f^{(2)}$, and $g^{(1)}$ to encode the same semantic concepts. For changed areas however, the encoded information in $f^{(1)}$ and $g^{(1)}$ should be similar, while $f^{(2)}$ should be dissimilar from the two. Based on this observation, we design a supervised contrastive loss that forces the network to learn better image features. More precisely, we apply a learnable projection $\pi$ to map $f^{(1)}$ and $f^{(2)}$ into the feature space of $g^{(1)}$. Then, we compute our contrastive loss as follows:
\begin{align}
    \mathcal{L}^{(contr)}_{(i,j)} &= -\text{sim}\left(g^{(1)}_{(i,j)}, \pi\big(f^{(1)}_{(i,j)}\big)\right) \\
    &+\begin{cases} 
        -\text{sim}\left(g^{(1)}_{(i,j)}, \pi\big(f^{(2)}_{(i,j)}\big)\right), & b_{(i,j)} = 0 \\
        \text{max}\left(\text{sim}\left(g^{(1)}_{(i,j)}, \pi\big(f^{(2)}_{(i,j)}\big)\right), 0 \right), & b_{(i,j)} = 1.
    \end{cases}\nonumber
\end{align}
Here, $sim(\cdot,\cdot)$ denotes the cosine similarity, and the loss is ultimately aggregated over all pixel locations $(i,j)$. Interestingly, we found it beneficial not to use a projection for $g^{(1)}$ and also not to backpropagate gradients coming from the contrastive loss through $g^{(1)}$. Details can be found in Section~\ref{sec:bcd}.

\subsection{Architecture Details}
\label{sec:arch-details}
Like~\cite{fhd}, we use the MixVisionTransformer from~\cite{segformer} as image encoder backbone. Due to the spatial attention in the encoder, the features $f^{(t)}_{(i,j)}$ are already contextualized, and a pointwise prediction head after a pointwise fusion module suffices to produce high-quality predictions. 
We leave SegFormer's prediction head unchanged and solely plug in our proposed feature fusion module between the backbone and the head. Thus, our module can be easily combined with other architectures.
Inside our feature fusion module, the merged features $(h_{k})_{k=1...K}$ are produced by $K$ two-layer MLPs, while the attention weights $a$ are predicted from $g^{(1)}$ via a linear layer.
To encode the semantic information $m^{(1)}$ into $g^{(1)}$, we use a three-layer CNN with a pointwise convolution followed by two convolutions with a kernel size of five and a dilation of two to increase the receptive field. When fusing the encoded map $g^{(1)}$, we use bilinear interpolation to comply with the different spatial resolutions of the different scales of the image feature maps.
The projection head $\pi$ for the contrastive loss is a pointwise MLP head in the style of SegFormer's prediction head. 
The overall training loss is the sum of our contrastive loss and the cross-entropy loss for the binary change map (as well as the semantic segmentation in SCD).

\section{Experiments}
\subsection{Datasets}
Since Conditional Change Detection requires semantic segmentation maps, which are not included in commonly used benchmark datasets such as LEVIR-CD~\cite{levir}, DSIFN-CD~\cite{dsifn}, and S2Looking~\cite{s2looking}, we conduct our experiments on DynamicEarthNet~\cite{dynearthnet} and HRSCD~\cite{hrscd}.
DynamicEarthNet is a recent dataset consisting of daily images of 75 areas of interest (AOIs) over a span of two years. Ground-truth annotations containing the seven land use classes "impervious surface", "agriculture", "forest \& other vegetation", " wetlands", "soil", "water", and "snow \& ice" are only available for the first day of each month and 55 AOIs. As our setting assumes semantic information as input, we only use the images with available annotations and selected 35, 10, and 10 AOIs for training, validation, and testing, respectively. Following~\cite{dynearthnet}, we omit the class "snow \& ice" from the evaluation because it only occurs in two AOIs. Targets for binary change detection were created by applying a pixelwise inequality on pairs of ground-truth semantic maps.

The High-Resolution Semantic Change Detection (HRSCD) dataset was proposed in~\cite{hrscd}. It contains 291 pairs of images, where the first was acquired in 2005/2006 and the second in 2012. Semantic ground-truth segmentation maps and binary change segmentation maps are available for all samples. The semantic classes are "artificial surface", "agricultural area", "forest", "wetland", and "water". We selected 191, 50, and 50 pairs for training, validation, and testing, respectively. Further details on both datasets can be found in the supplementary material.

\subsection{Baselines}
As baselines derived from semantic segmentation models, we employ SegFormer~\cite{segformer}, Swin + UPerNet~\cite{swin, upernet}, and SETR-PUP~\cite{setr} for change detection by taking the pixelwise inequality of the predicted semantic segmentations $\hat{m}^{(1)}$ and $\hat{m}^{(2)}$ as binary change prediction.
The concatenation baseline for bi-temporal BCD resembles the SegFormer architecture, but concatenates the image features $f^{(1)}$ and $f^{(2)}$ before passing them to a head that directly predicts the binary change map.
These baselines can be easily reused for Conditional BCD. For the semantic segmentation baselines, we apply the pixelwise inequality to $m^{(1)}$ and $\hat{m}^{(2)}$ instead of $\hat{m}^{(1)}$ and $\hat{m}^{(2)}$. For the concatenation baseline, we concatenate map features $g^{(1)}$ extracted by a map encoder as described in Section~\ref{sec:arch-details} to the image features $f^{(1)}$ and $f^{(2)}$. 
As state-of-the-art methods for bi-temporal BCD, we select FHD~\cite{fhd}, ChangerEx~\cite{changer}, and ChangeFormer~\cite{changeformer} as they are recent and have been shown to perform best on the common benchmarks. We also extend these methods for Conditional BCD by concatenating map features to merged bi-temporal image features. 
Further, we consider another baseline for Conditional BCD, which has a separate concatenation-based BCD head for every semantic class such that the final change prediction can be taken from the head corresponding to the pre-change semantic class for every pixel.
To have a strong competitor for our feature fusion module, we also use a pixel-wise Dynamic MLP-C~\cite{dynamicmlp} for fusing semantic input with image features. 
For training details as well as a comparison of model sizes, we refer to the supplementary material.


\subsection{Binary Change Detection}
\label{sec:bcd} 
\begin{table}
\centering
\begin{tabular}{@{\hspace{0em}}c@{\hspace{0.2em}}|c@{\hspace{1em}}c|c@{\hspace{1em}}c}
\hline
\multirow{2}{*}{\textbf{Method}} & \multicolumn{2}{c|}{\textbf{DynENet}} & \multicolumn{2}{c}{\textbf{HRSCD}} \\
  & \textbf{F1} & \textbf{IoU} & \textbf{F1} & \textbf{IoU} \\ \hline
\textbf{\textit{\underline{Bi-temporal BCD:}}} &  &  &  & \\
SegFormer~\cite{segformer}       & 20.3 & 11.3 & 17.8 & \enskip 9.7 \\
Swin + UPerNet~\cite{swin,upernet} & 20.1 & 11.1 & 17.6 & \enskip 9.7 \\
SETR-PUP~\cite{setr}             & 19.4 & 10.8 & 14.8 & \enskip 8.0 \\
Concatenation                    & 20.7 & 11.6 & 41.2 & 25.9 \\
FHD~\cite{fhd}                   & 17.2 & \enskip 9.4 & 45.2 & 29.2 \\ 
ChangerEx~\cite{changer}         & 21.1 & 11.8 & 37.0 & 22.7 \\
ChangeFormer~\cite{changeformer} & 20.7 & 11.5 & \underline{45.7} & \underline{29.6} \\
\textbf{MapFormer}$_{K=10}$\,(w/o~$m^{(1)}$)     & \underline{28.6} & \underline{16.7} & 43.9 & 28.2 \\
\hline
\textbf{\textit{\underline{Conditional BCD:}}} &  &  &  & \\ 
SegFormer~\cite{segformer}       &  21.2  &  11.9   & \enskip 5.6 & 2.8 \\
Swin + UPerNet~\cite{swin,upernet} & 21.8 & 12.2 & \enskip 5.6 & 2.8 \\
SETR-PUP~\cite{setr}             & 20.8 & 11.6 & \enskip 5.4 & 2.8 \\
Concatenation                    &  33.5  &  20.1   & 62.1 & 45.1 \\   
FHD~\cite{fhd} + Concat.                   & 33.2 & 19.9 & 61.2 & 44.1 \\ 
ChangerEx~\cite{changer} + Concat.         & 30.9 & 18.3 & 58.1 & 41.0 \\ 
ChangeFormer~\cite{changeformer} + Concat. & 31.4 & 18.6 & 61.5 & 44.4 \\
Classwise CD-Heads                         & 37.0 & 22.7 & 62.3 & 45.3 \\ 
DynamicMLP~\cite{dynamicmlp}     &  34.5  &  20.8   & 62.7 & 45.7 \\
\textbf{MapFormer}$_{K=5}$ & 34.8 & 21.1 & \textbf{64.9} & \textbf{48.0} \\
\textbf{MapFormer}$_{K=10}$                    & \textbf{38.0}   & \textbf{23.5}    & 64.5 & 47.7 \\
\textbf{MapFormer}$_{K=15}$& 36.4 & 22.2 & 64.5 & 47.6 \\[5pt]
\underline{\textit{Ablations} (MapFormer$_{K=10}$)}: &  &  &  & \\ 
w/o~constrastive  &  34.5  &  20.9   & 63.8 &  46.8 \\
w/o~stop-gradient &  36.7  & 22.5    & 62.5 &  45.4 \\
w/~map proj.       &  35.8  & 21.8    & 60.7 &  45.9 \\
high-level\,info& 29.8 & 17.6 & 64.4 & 47.5 \\
low-res.\,info  & 30.4 & 17.9 & 60.3 & 43.2 \\
\hline
\end{tabular}
\caption{(\textbf{Conditional) Binary Change Detection} as well as ablations. Our method performs best and is still highly competitive without ground-truth semantic input.
}
\label{tab:bc}
\end{table}

\textbf{Overall Performance}
In BCD, there is only a single objective (binary change IoU); thus, our approach can be best compared to other methods.
In Table~\ref{tab:bc}, we present BCD results for baselines, state-of-the-art methods, and our method. 
It is evident that Conditional BCD greatly outperforms all existing bi-temporal methods -- even with the simple concatenation baseline (11.8\% vs.~20.1\% and 29.6\% vs.~45.1\% IoU, resp.). MapFormer further pushes the performance on both datasets to 23.5\% and 48.0\% IoU, respectively. Particularly, our method also outperforms DynamicMLP~\cite{dynamicmlp}, a state-of-the-art method for multi-modal feature fusion. This holds for all tested values of our only hyperparameter $K$ and even without our contrastive loss (see Ablations). 
Remarkably, when MapFormer is applied without the true map $m^{(1)}$, but only predicted segmentation maps $\hat{m}^{(1)}$ (w/o $m^{(1)}$), the performance is still highly competitive, even achieving by far the best performance among the bi-temporal methods on DynamicEarthNet with an IoU of 16.7\%.
Furthermore, it is noteworthy that the semantic segmentation baselines SegFormer, Swin + UPerNet and SETR perform poorly in Conditional BCD, especially on HRSCD. We attribute this to the fact that every misclassified pixel leads to a false positive change prediction in this setup, causing a very low precision. In bi-temporal BCD, identical classification errors in $\hat{m}^{(1)}$ and $\hat{m}^{(2)}$ may still lead to true negative change predictions.

\textbf{Ablations}
\begin{table*}
    \begin{minipage}{0.34\linewidth}
            \begin{tabular}{@{\hspace{0.2em}}c@{\hspace{0.2em}}|@{\hspace{0.3em}}c@{\hspace{0.6em}}c@{\hspace{0.3em}}|@{\hspace{0.6em}}c@{\hspace{0.6em}}c@{\hspace{0.3em}}}
            \hline
            \multirow{2}{*}{\textbf{Method}} & \multicolumn{2}{@{\hspace{0.2em}}c@{\hspace{0.6em}}|@{\hspace{0.6em}}}{\textbf{DynENet}} & \multicolumn{2}{@{\hspace{0.3em}}c@{\hspace{0.3em}}}{\textbf{HRSCD}} \\
              & \textbf{F1} & \textbf{IoU} & \textbf{F1} & \textbf{IoU} \\ \hline
            SegFormer       &  21.2  &  11.9   & \enskip 5.6 & \enskip 2.8 \\
            Swin+UPerNet    &  21.8  &  12.2   & \enskip 5.6 & \enskip 2.8 \\
            SETR-PUP        &  20.8  &  11.6   & \enskip 5.4 & \enskip 2.8 \\
            Concatenation  & 31.8 & 18.9 & 61.5 & 44.4 \\ 
            MapFormer & \textbf{32.0} & \textbf{19.0} & \textbf{62.1} & \textbf{45.0} \\
            \hline
            \end{tabular}
            \caption{\textbf{Cross-modal Binary Change Detection.} MapFormer and the concatenation baseline outperform SegFormer.}
            \label{tab:cross-modal-bc}
    \end{minipage} 
    \hfill
    \begin{minipage}{.65\linewidth}
        \hfill
        \begin{tabular}{@{\hspace{0.1em}}c@{\hspace{0.2em}}|@{\hspace{0.2em}}c@{\hspace{0.7em}}c@{\hspace{0.7em}}c@{\hspace{0.3em}}c@{\hspace{0.1em}}|@{\hspace{0.2em}}c@{\hspace{0.7em}}c@{\hspace{0.7em}}c@{\hspace{0.3em}}c@{\hspace{0.1em}}}
            \hline
            \multirow{2}{*}{\textbf{Method}} & \multicolumn{4}{c|@{\hspace{0.2em}}}{\textbf{DynamicEarthNet}} & \multicolumn{4}{c}{\textbf{HRSCD}} \\
             & \textbf{BC} & \textbf{SC} & \textbf{SCS} & \textbf{mIoU} & \textbf{BC} & \textbf{SC} & \textbf{SCS} & \textbf{mIoU} \\ \hline
            SegFormer      & 11.3 & \textbf{31.5} & 21.4 & 39.7 & \enskip 9.7 & 20.3 & 15.0 & 52.7 \\
            Swin + UperNet & 11.1 & 28.0 & 19.6 & 41.4 & \enskip 9.7 & 20.9 & 15.3 & 53.1 \\
            SETR-PUP       & 10.8 & 26.3 & 18.5 & 38.5 & \enskip 8.5 & \textbf{21.7} & 14.9 & 50.5 \\
            MapFormer\,(sem.\,seg.\,on\,$f^{(2)}$) & 23.0 & 29.0 & \textbf{26.0} & 39.9 & 48.0 & 21.5 & \textbf{34.7} & 52.9 \\
            MapFormer\,(sem.\,seg.\,on\,$f$)       & \textbf{23.1} & 13.1 & 18.1 & 61.5 & \textbf{48.2} & 12.5 & 30.4 & \textbf{72.3} \\
            MapFormer\,(Cross-modal)         & 20.4 & 30.7 & 25.5 & \textbf{40.3} & 45.6 & \textbf{21.7} & 33.5 & 49.9 \\
            \hline
        \end{tabular}
        \caption{\textbf{(Conditional) Semantic Change Detection.} Predicting $\hat{m}^{(2)}$ based on $f^{(2)}$ yields the best results w.r.t.~the main objective SCS.}
        \label{tab:scd}
    \end{minipage} 
\end{table*}
\begin{figure*}[t]
    \centering
    \begin{subfigure}{0.92\textwidth}
        \centering
        \begin{subfigure}{0.2\columnwidth}
            \subcaption*{$I^{(1)}$}
            \includegraphics[width=\columnwidth]{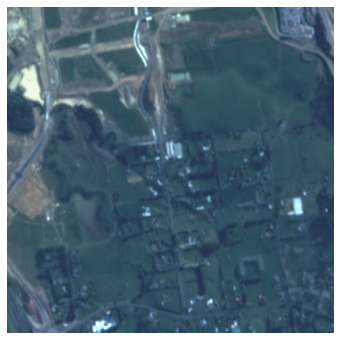}
        \end{subfigure} \hspace{-0.88em}
        \begin{subfigure}{0.2\columnwidth}
            \subcaption*{$I^{(2)}$}
            \includegraphics[width=\columnwidth]{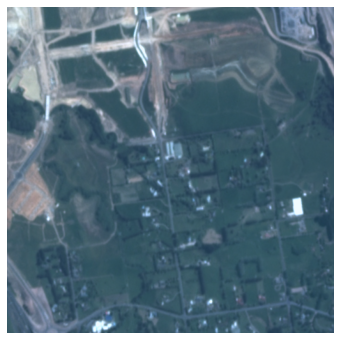}
        \end{subfigure} \hspace{-0.88em}
        \begin{subfigure}{0.2\columnwidth}
            \subcaption*{$m^{(1)}$}
            \includegraphics[width=\columnwidth]{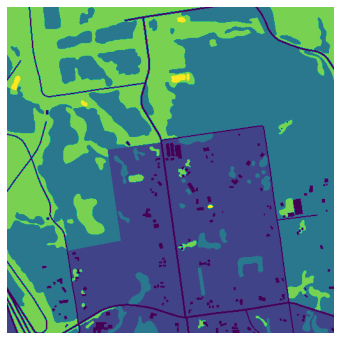}
        \end{subfigure} \hspace{-0.88em}
        \begin{subfigure}{0.2\columnwidth}
            \subcaption*{$m^{(2)}$}
            \includegraphics[width=\columnwidth]{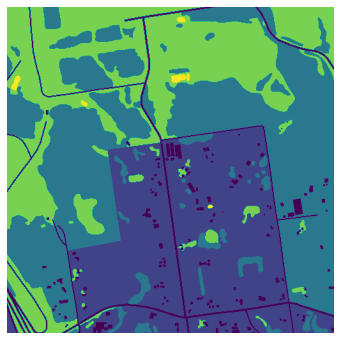}
        \end{subfigure} \hspace{-0.88em}
        \begin{subfigure}{0.2\columnwidth}
            \subcaption*{ Ground truth $b$}
            \includegraphics[width=\columnwidth]{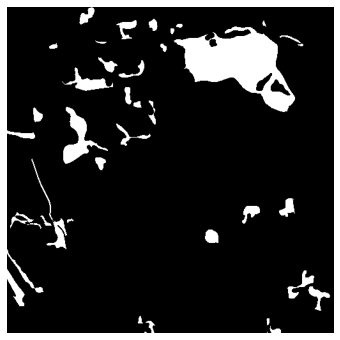}
        \end{subfigure} 
        \vspace{-0.5em}
        
        \begin{subfigure}{0.2\columnwidth}
            \includegraphics[width=\columnwidth]{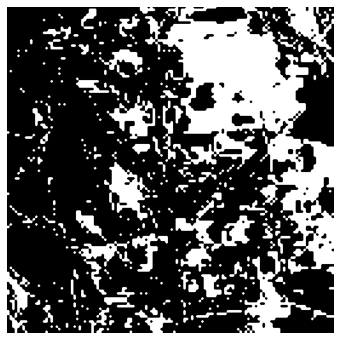}
            \subcaption*{SegFormer}
        \end{subfigure} \hspace{-0.88em}
        \begin{subfigure}{0.2\columnwidth}
            \includegraphics[width=\columnwidth]{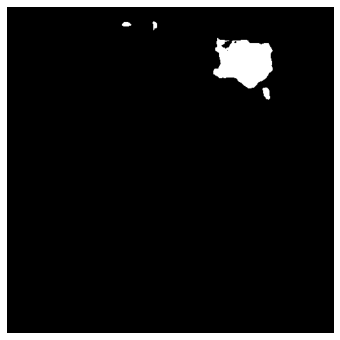}
            \subcaption*{FHD}
        \end{subfigure} \hspace{-0.88em}
        \begin{subfigure}{0.2\columnwidth}
            \includegraphics[width=\columnwidth]{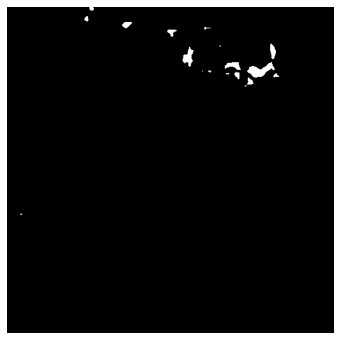}
            \subcaption*{ChangerEx}
        \end{subfigure} \hspace{-0.88em}
        \begin{subfigure}{0.2\columnwidth}
            \includegraphics[width=\columnwidth]{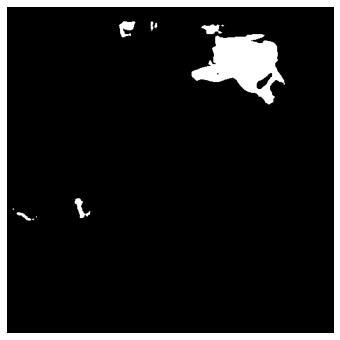}
            \subcaption*{ChangeFormer}
        \end{subfigure} \hspace{-0.88em}
        \begin{subfigure}{0.2\columnwidth}
            \includegraphics[width=\columnwidth]{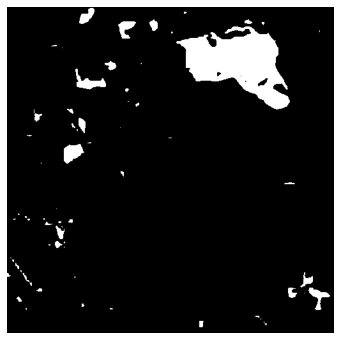}
            \subcaption*{MapFormer}
        \end{subfigure} \hspace{0em}
        \subcaption{Binary change prediction for different methods.}
    \end{subfigure}
    
    \begin{subfigure}{0.92\textwidth}
        \centering
        \begin{subfigure}{0.171\columnwidth}
            \includegraphics[width=\columnwidth]{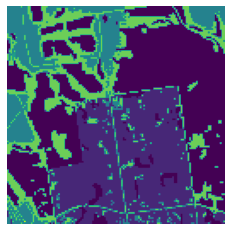}
        \end{subfigure}\hspace{-0.8em}
        \begin{subfigure}{0.171\columnwidth}
            \includegraphics[width=\columnwidth]{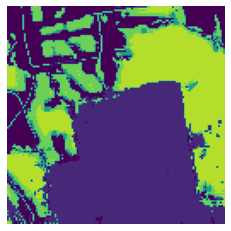}
        \end{subfigure}\hspace{-0.8em}
        \begin{subfigure}{0.171\columnwidth}
            \includegraphics[width=\columnwidth]{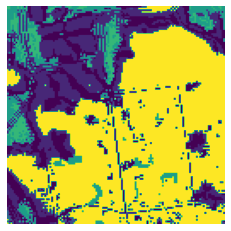}
        \end{subfigure}\hspace{-0.8em}
        \begin{subfigure}{0.171\columnwidth}
            \includegraphics[width=\columnwidth]{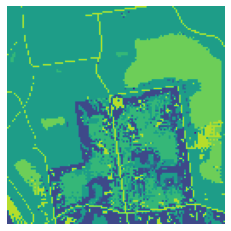}
        \end{subfigure}\hspace{-0.8em}
        \begin{subfigure}{0.171\columnwidth}
            \includegraphics[width=\columnwidth]{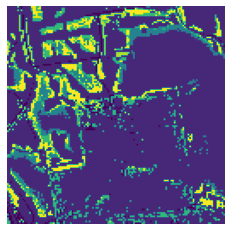}
        \end{subfigure}\hspace{-0.8em}
        \begin{subfigure}{0.171\columnwidth}
            \includegraphics[width=\columnwidth]{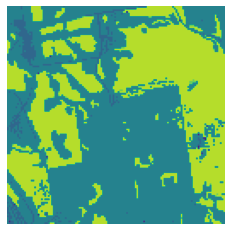}
        \end{subfigure} \hspace{0em}
    \subcaption{Visualization of attentions weights. Colors encode which of the representations $(h_k)_{k=1...K}$ received most attention for six selected channels $d$, exhibiting a focus on different aspects of the features for different $k$ and $d$.}
    \end{subfigure}

    \begin{subfigure}{0.98\textwidth}
        \centering
        \begin{subfigure}{0.25\textwidth}
            \centering
            \includegraphics[width=0.7\columnwidth]{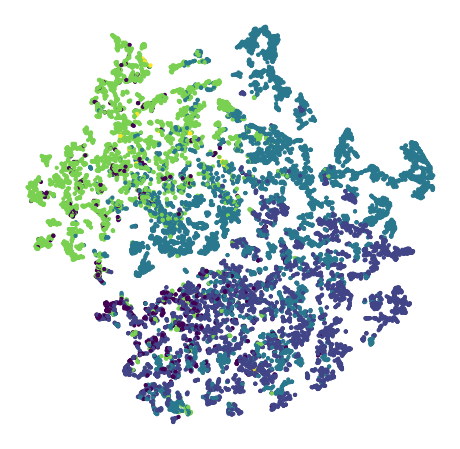}
            \subcaption*{(c.1) FHD [sem.~classes]}
        \end{subfigure}\hspace{-0.88em}
        \begin{subfigure}{0.25\textwidth}
            \centering
            \includegraphics[width=0.7\columnwidth]{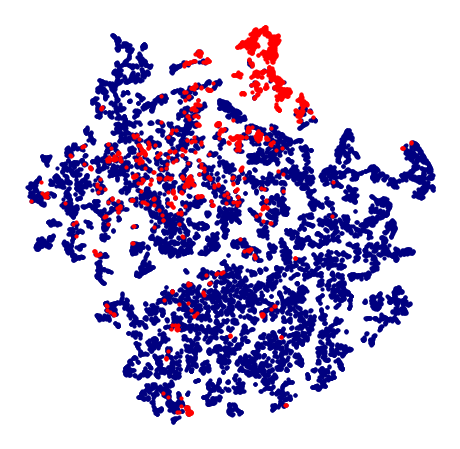}
            \subcaption*{(c.2) FHD [bin.~change]}
        \end{subfigure}\hspace{-0.88em}
        \begin{subfigure}{0.25\textwidth}
            \centering
            \includegraphics[width=0.7\columnwidth]{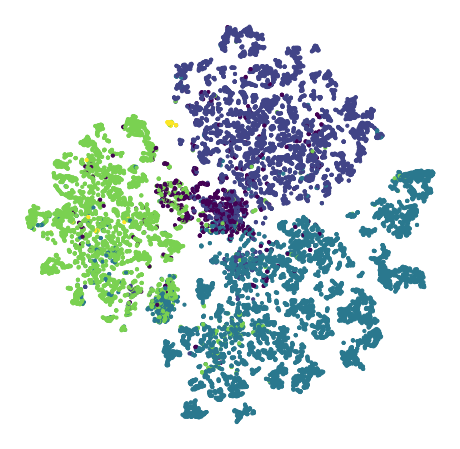}
            \subcaption*{(c.3) MapFormer [sem.~classes]}
        \end{subfigure}\hspace{-0.88em}
        \begin{subfigure}{0.25\textwidth}
            \centering
            \includegraphics[width=0.7\columnwidth]{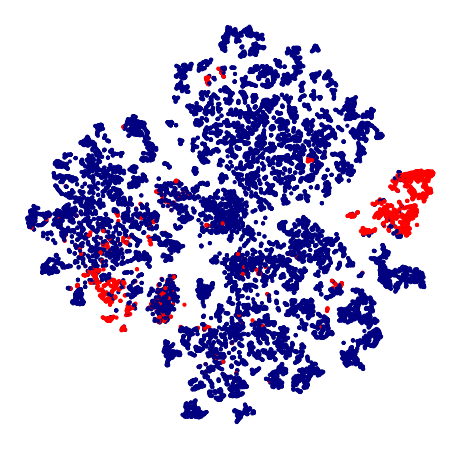} 
            \subcaption*{(c.4) MapFormer [bin.~change]}
        \end{subfigure}
    \subcaption{t-SNE~\cite{t-sne} visualizations of hidden features of FHD and our model. In the plots (c.1) and (c.3), colors correspond to the semantic classes of $m1$, whereas in (c.2) and (c.4), colors encode the binary change ground truth $b$ (\textcolor{red}{change}, \textcolor{blue}{no change}).}
    \end{subfigure}
    
    \caption{Qualitative results and insights for a sample of DynamicEarthNet.}
    \label{fig:qualitative-results}
\end{figure*}

In the lower part of Table~\ref{tab:bc}, we present several ablation experiments to analyze MapFormer's contributing factors and robustness.
First, we study the effect of our design choices in the contrastive loss module. When the contrastive loss is completely omitted (w/o contrastive), performance drops on both datasets by 2.6\% and 0.9\% IoU, respectively. Furthermore, we compare MapFormer with modifications where we omitted the stop-gradient operation on the map features $g^{(1)}$ (w/o stop-gradient) and a version where we included a linear projection layer to map $g^{(1)}$ to the joint embedding space with $\pi\big( f^{(t)}\big)$ (w/ map proj.). Both variants do not reach the performance of our proposed method. Thus, we conclude that gradients coming from the contrastive loss flowing through the map features $g^{(1)}$ have a detrimental effect on their representation.

Finally, we investigate the effect of the quality of the semantic information $m^{(1)}$. To this end, we merge some classes in the pre-change maps to simulate that only high-level semantic information is available (high-level info). For DynamicEarthNet, we merge the classes "agriculture", "forest \& other vegetation" and "soil" as well as "wetlands", "water" and "snow \& ice" into two high-level classes, respectively. For HRSCD, we solely distinguish between "artificial" and "rest". Apparently, this high-level information still suffices to generate better change predictions than the bi-temporal change detection methods. However, we observe a significant performance drop on DynamicEarthNet (5.9\% IoU) , while the performance only degrades marginally on HRSCD (0.2\% IoU).
Furthermore, we simulate low-resolution semantic information by downsampling with a factor of 32 (low-res.~info). Here, we can see that the performance of our method degrades with the low-resolution input (by 5.6\% and 4.8\% IoU, resp.).  Nevertheless, it is still clearly superior to the bi-temporal change detection methods. In conclusion, our results show that the semantic input quality correlates to the Conditional Change Detection performance. However, low-quality input still suffices to achieve significantly better results than methods not relying on this kind of information. 
Further ablations can be found in the supplementary material.

\subsection{Cross-modal Binary Change Detection}
\label{sec:cross-modal-bcd}
We also evaluate MapFormer on the task of Cross-modal BCD, i.e. change detection when only post-change imagery and pre-change map information are given (see Table~\ref{tab:cross-modal-bc}). Here, the concatenation baseline omits the pre-change image features $f^{(1)}$. The semantic segmentation baselines are identical to the ones for Conditional BCD in Table~\ref{tab:bc} as they do not utilize $\hat{m}^{(1)}$. Again, we can see that MapFormer performs best, although the margin to the concatenation baseline is rather small (0.1\% and 0.6\% IoU, resp.). However, comparing the results with those in Table~\ref{tab:bc}, we observe that bi-temporal methods are greatly outperformed in this setting as well, and the performance is still relatively close to Conditional BCD, which relies on bi-temporal images. 
This highlights the value of semantic input for change detection and shows that bi-temporal images are not critical for change detection when pre-change maps are available.

\subsection{Semantic Change Detection}
\label{sec:scd}
For semantic change detection, one can integrate additional semantic segmentation heads into our framework. We provide the results for Conditional SCD and Cross-modal SCD in Table~\ref{tab:scd}. Here, we follow the notation of~\cite{dynearthnet} and denote the binary change IoU with BC (instead of IoU in Tables~\ref{tab:bc},\ref{tab:cross-modal-bc}).
In the first three rows of Table~\ref{tab:scd}, we reuse the semantic segmentation baselines of Table~\ref{tab:bc}. 
The fourth and fifth rows correspond to MapFormer, with the semantic segmentation outputs being predicted from the uni-temporal image features (sem.~seg.~on $f^{(2)}$) and the merged features (sem.~seg.~on $f$), respectively. Apparently, the version with a segmentation head applied on $f^{(2)}$ performs better w.r.t.~BC, SC, and SCS, whereas the joint version is superior w.r.t.~mIoU. Our reasoning is that the imbalance between changed and unchanged areas in the data causes the model using $f$ to tend to repeat the given semantic information, which is not accessible through $f^{(2)}$. This hurts the performance on changed pixels (SC), but leads to higher overall mIoU.
In the last row of Table~\ref{tab:scd}, we show the results for MapFormer solely based on uni-temporal images (Cross-modal), using $f^{(2)}$ for semantic segmentation. Similar to our observations for BCD, this model is competitive to its counterpart for Conditional SCD (25.5\% vs.~26.0\% and 33.5\% vs.~34.7\% SCS, resp.). This implies that $m^{(1)}$ has a stronger impact on the performance than $I^{(1)}$.

\subsection{Qualitative Results}
\label{sec:qualitative}

The first row of Figure~\ref{fig:qualitative-results}~(a) depicts inputs and ground-truth targets for a sample of DynamicEarthNet. 
The second row shows the binary change predictions of several methods for this sample. We can see that SegFormer suffers from many false positive predictions, while the other state-of-the-art methods struggle to detect the changed areas. In contrast, MapFormer produces a convincing change mask where only small details are missing.

In Figure~\ref{fig:qualitative-results}~(b), we visualize the attention weights of MapFormer for this sample. The visualizations correspond to the argmax over the $K$-dimension of the attention weights $a_{(\cdot,\cdot,k,d)}$ for six different channels $d$. In other words, the colors in each visualization show which of the features $(h_k)_{k=1...K}$ receive most attention for the selected channels. Comparing them, we can see that the maps focus on different aspects of the semantic information $m_1$. 
For example, the first attention map mostly resembles the original map $m^{(1)}$, whereas the second and third visualizations resemble more coarse versions of $m_1$. On the other hand, the fourth map focuses more on small and thin segments such as roads. The remaining two attention maps seem to aim at separating the features for specific semantic classes.

To provide further insight, we compare t-SNE~\cite{t-sne} visualizations of the hidden features of MapFormer and FHD in Figure~\ref{fig:qualitative-results}~(c). Each point in the visualization represents the hidden feature for one pixel in the sample depicted in subfigure (a). For MapFormer, the feature space is much less entangled, and the semantic classes of the pixels and the binary change ground truth can be separated from other pixels relatively well. In contrast, the FHD representations for different semantic and binary change classes are highly mixed, leading to prediction errors. 

\section{Conclusion}
In this paper, we have introduced Conditional Change Detection, a new paradigm for change detection. We have shown that using available pre-change semantic information as input leads to significant improvements over existing methods for change detection. This holds even for a simple baseline that solely relies on feature concatenation. To further improve the performance, we have proposed MapFormer, a novel architecture outperforming its competitors in all considered settings.
A limitation of our approach is the requirement of semantic pre-change information. 
In the future, we plan to mitigate this dependency by leveraging pre-trained models to generate pre-change maps via transfer learning.
Further, we see room for improvement w.r.t.~the semantic segmentation pipeline of our approach if more sophisticated techniques are integrated for this purpose. 
We hope this work will lead to further research following the avenue of Conditional Change Detection. In particular, we believe that more diverse datasets, more sophisticated methods, and other settings, such as uni-temporal supervision as in~\cite{change-is-everywhere}, will greatly benefit this line of research. 

{\small
\bibliographystyle{iccv2023AuthorKit/ieee_fullname}

}

\end{document}